\documentclass[11pt, a4paper, logo, onecolumn, nonumbering]{googledeepmind}

\pdftrailerid{redacted}

\makeatletter
\renewcommand\bibentry[1]{\nocite{#1}{\frenchspacing\@nameuse{BR@r@#1\@extra@b@citeb}}}
\makeatother

\usepackage{kantlipsum, lipsum}
\usepackage{dsfont}
\usepackage{gdm-colors}

\usepackage[authoryear, sort&compress, round]{natbib}

\graphicspath{{figures/}}

\title{Societal and technological progress as sewing an ever-growing, ever-changing, patchy, and polychrome quilt}

\correspondingauthor{Joel Z. Leibo (jzl@google.com)}

\reportnumber{} 

\author[1]{Joel Z.~Leibo}
\author[1]{Alexander Sasha Vezhnevets}
\author[1, 2]{William A. Cunningham}
\author[1]{S\'{e}bastien Krier}
\author[3]{Manfred Diaz}
\author[1]{Simon Osindero}

\affil[1]{Google DeepMind}
\affil[2]{University of Toronto}
\affil[3]{Mila - Qu\'{e}bec AI Institute}

\begin{abstract}
Artificial Intelligence (AI) systems are increasingly placed in positions where their decisions have real consequences, e.g.,~moderating online spaces, conducting research, and advising on policy. Ensuring they operate in a safe and ethically acceptable fashion is thus critical. However, most solutions have been a form of one-size-fits-all ``alignment''. We are worried that such systems, which overlook enduring moral diversity, will spark resistance, erode trust, and destabilize our institutions. This paper traces the underlying problem to an often-unstated Axiom of Rational Convergence: the idea that under ideal conditions, rational agents will converge in the limit of conversation on a single ethics. Treating that premise as both optional and doubtful, we propose what we call the appropriateness framework: an alternative approach grounded in conflict theory, cultural evolution, multi-agent systems, and institutional economics. The appropriateness framework treats persistent disagreement as the normal case and designs for it by applying four principles: (1) contextual grounding, (2) community customization, (3) continual adaptation, and (4) polycentric governance. We argue here that adopting these design principles is a good way to shift the main alignment metaphor from moral unification to a more productive metaphor of conflict management, and that taking this step is both desirable and urgent.
\end{abstract}

\begin{document}

\maketitle

Disclaimer: These are our own opinions; they do not represent the views of Google DeepMind as a whole or its broader community of safety researchers.

\section*{The Patchwork Quilt of Human Coexistence}

\noindent\fbox{%
    \parbox{0.95\textwidth}{%
``We pragmatists think of moral progress as more like sewing together a very large, elaborate, polychrome quilt, than like getting a clearer vision of something true and deep.''
\flushright{---\cite{rorty2021pragmatism} pg.~141}
  }%
}

Quite a lot of thinking in AI alignment, particularly within the rationalist tradition associated with digital communities such as LessWrong, implicitly or explicitly appears to rest upon something we might call an `Axiom of Rational Convergence' (e.g.~\cite{yudkowsky2004coherent}). This is the powerful idea that under sufficiently ideal epistemic conditions---ample time, information, reasoning ability, freedom from bias or coercion---rational agents will ultimately converge on a single, correct set of beliefs, values, or plans, effectively identifying ``the truth''. We suspect that the axiom of rational convergence is independent of the rest of AI safety and ethics. By this, we mean something analogous to the way different geometries can be constructed by selecting different ``parallel axioms'' e.g. Euclid's parallel axiom gives you the geometry of a plane, postulating instead that parallel lines converge gives you elliptic geometry, and postulating that parallel lines diverge generates hyperbolic geometry. The choice of which parallel axiom to adopt is like the choice of what to study (e.g. a plane or a sphere). Just as selecting different parallel axioms generates distinct yet internally consistent geometries, the choice of whether to accept the Axiom of Rational Convergence shapes our theories about AI alignment and ethics. Could the Axiom of Rational Convergence be similarly independent of other foundational commitments? Its appeal lies in its offering a potential objective grounding for AI goals, ``something true and deep'' towards which alignment can aim. However, what if this axiom doesn't hold, or what if we simply choose not to adopt it as the basis for navigating the complexities of advanced AI? Here we explore the consequences of constructing an approach to AI safety that rejects the axiom of rational convergence. We will try to construct a framework that takes disagreements between individuals as basic and persisting indefinitely, not as mere pitstops on the way to rational convergence.

Does rejecting the Axiom of Rational Convergence mean denying a distinction between fact and opinion? In important ways, yes. The cognitive opacity of moral categories makes this distinction less useful than it initially appears. While growing up, we internalize both universally shared ``thin'' moral principles and culturally contingent ``thick'' ethical systems \citep{walzer1994thick}; but crucially, for any given query, and using just our own culture's vocabulary, we cannot distinguish whether our answer relies on principles that other cultures may also accept versus principles unique to our own culture \citep{rorty1989contingency}. How can we tell if a particular cherished rule of social interaction is really critical for cooperation given that transgression is met with the same social punishment regardless? We assume it is generally difficult to reliably distinguish important rules from ``silly'' rules \citep{hadfield2019legible, koster2022spurious}. Therefore, where the ``default rationalist view'' would typically assume rational convergence for ``fact-like'' questions but allow divergence on ``opinion-like'' questions, we instead propose to eschew assuming convergence for any kind of question. Propositions we would ordinarily be tempted to call ``facts'' are, in our view, governed by \textit{epistemic norms} that control justificatory standards and appropriate evidence integration, and which crucially are just as culturally contingent as any other kind of norm. We understand this core assumption of our framework to be empirically justified in light of research on causally opaque cultural knowledge showing that humans depend on socially transmitted information without necessarily understanding the causal mechanisms that determine the effectiveness of any particular cultural element (e.g.~\cite{boyd2011cultural, derex2019causal, henrich2021cultural, harris2021role, derex2025social}). If such causal opacity dominates even in the relatively simple domain of tasks necessary for survival as a subsistence forager, then it is surely also present in the domain of norm cognition that governs cooperation in complex modern societies\footnote{We are aware of older empirical work that diverges from this perspective such as that of \cite{turiel1983development} which saw moral principles of post-enlightenment liberalism (e.g.~the focus on harm minimization and equality maximization) as a universal feature arising in individual moral development. We consider this model as already having been effectively rebutted by others who showed it fails to account for cross cultural data; see review by \cite{machery2022moral}. So we need not say any more about it here.}, see also \citep{whitehouse2021ritual, jagiello2022tradition}. We interpret this to suggest that culture evolves, at least in part, by mechanisms other than the kind of rational conversation imagined in the Axiom of Rational Convergence\footnote{In this light, theoretical models that envision a process of deliberation on norms without suggesting the convergence of said process are especially suggestive e.g.~\cite{heyes2022rethinking}.}.

There is always a sense of arbitrariness in a choice of axioms. But it's not an arbitrariness without consequence. Clearly, the choice has large implications for the usefulness of the resulting theory in particular domains. A theory of planar geometry is no good for describing geometry on the surface of a sphere. We think, for AI alignment, choices concerning the axiom of rational convergence are just as consequential, or even more so. Indeed, rejecting the Axiom of Rational Convergence requires a different guiding image for social progress. We find one in Richard Rorty's metaphor (which anchors this article): instead of striving for a ``clearer vision of something true and deep''---a singular, universal `human volition'---social progress instead resembles ``sewing together a very large, elaborate, polychrome quilt''. This view sees human societies not as converging towards uniformity, but functioning as intricate patchworks built from diverse communities with persistently divergent values, norms, and worldviews, held together by the stitches of social conventions, institutions, and negotiation \citep{mouffe1999deliberative, march2011logic}. As AI systems increasingly speak and act within our social worlds, they too must navigate this patchwork quilt of contextual norms. Their failures manifest not as ``misalignment'' with an abstract ideal, but rather as context-inappropriate behavior---acting like a misplaced thread or a clashing pattern, violating the explicit and tacit rules governing a specific ``patch'' of the quilt. Grounding AI safety in this way leads to what we call the \textbf{appropriateness} framework \citep{leibo2024theory}.

This perspective is important because powerful AI systems exhibiting context-inappropriate behavior can generate significant social instability, erode public trust, and provoke counterproductive regulatory responses. Such instability could hinder the careful, coordinated global effort that will be required to manage the development of transformative AI safely. Even assuming a best effort to apply technical alignment techniques before building powerful AI, it is well acknowledged that, with superintelligence, there will always remain substantial residual risks stemming from the ``specification gap''---unavoidable differences between the complex, nuanced, context-dependent goals and values that humans actually have and the necessarily simplified objective function, reward signal, or behavioral rule that we can encode into AI \citep{amodei2016concrete, hadfield2019incomplete}. The approach we develop here takes this residual misalignment as given and asks how we can prevent it from engendering violent social instability. By focusing on preventing these destabilizing social dynamics, we aim to safeguard the stable societal conditions necessary for careful development and governance of advanced AI. We don't think this approach diminishes concerns around power-seeking AI systems, but it does reconceptualize them---unifying them with general concerns around concentration of power in human groups (and human-AI coalitions).

We proceed from the core assumption that stable human coexistence (a precondition for flourishing), particularly in diverse societies, is made possible not by achieving rational convergence on values, but by relying on practical social technologies---like conventions, norms, and institutions---to manage conflict and enable coordination \citep{north1990institutions, ostrom2009understanding, hadfield2012law}. We do not see persistent disagreement as problematic or puzzling. For the purpose of effectively navigating coexistence, we choose to treat the  disagreements as basic elements to work with and proceed as if fundamental differences are enduring features of our shared landscape. We will argue that this approach is more practical than alternatives which view disagreements as mere errors on a path to rational convergence. For example, Christianity posits that a human being possesses a unique, atomic (indivisible) essence---a soul, while Buddhism denies its existence (anattā, one of four noble truths), they differ in basic ontology. Cross-cultural divergences manifest in tangible, everyday practices. Consider differing cultural conceptions of fairness: is it strictly equal distribution (leading perhaps to rotational systems for resource access), allocation based on need (requiring complex social mechanisms for assessment), reward proportional to effort (underpinning certain market ideologies), or upholding established social hierarchies (reflected in caste systems or traditional aristocratic privileges)? Each view implies vastly different economic arrangements, educational priorities, and dispute resolution practices, shaping who gets what and why \citep{henrich2005economic}. Or consider the norms of appropriate communication: in some contexts, blunt directness is valued as honesty; in others, elaborate indirectness, careful face-saving, and attention to relational hierarchy are paramount \citep{fiske1992four, wardhaugh2021introduction, henrich2020weirdest, nisbett2003geography}. An AI trained exclusively on the norms and reasoning styles of one social context will inevitably appear rude, illogical, obtuse, or even hostile when interacting in a different context.

These persistently divergent norms, lifeways, and foundational beliefs challenge the notion that disagreements are merely temporary states awaiting rational convergence \citep{rorty1989contingency}. We think that whether the arc of humanity's collective conversation ultimately bends toward Truth is not something to assume and derive further consequences from, but rather is better seen as something to study empirically. And, empirically speaking, there really do seem to be persistent disagreements that don't go away with more education, information, time to think, freedom from domination, etc, or any of the other hypothesized impediments to Truth's emergence from collective conversation \citep{graham2009liberals, taber2016illusion, iyengar2019scientific}. Therefore rather than minimizing disagreement, we propose to focus on resolving conflict.

Think of society as a patchwork quilt composed of diverse communities, each with its own internal norms and expectations. Within these communities---e.g.~trade unions enforcing strike discipline or religious groups defining attendance practices---members understand the specific rules and the social consequences for deviating \citep{ullmann1977emergence, ostrom1990governing, bicchieri2005grammar}. These internal dynamics shape behavior within each distinct patch of the quilt, fostering cohesion through shared, localized understandings of appropriate conduct.

Crucially, these community-specific norms often extend outwards, influencing interactions with those outside the immediate group and sometimes leading to friction. A religious bakery owner might refuse to bake a cake for a same-sex wedding, creating conflict at the boundary between different patches. While such actions spark disagreement (and we deliberately choose contested examples), a robust social fabric must often find ways to manage these kinds of boundary-crossing claims rather than simply rejecting them \citep{merry1988legal}. This isn't unique to religion; consider how environmental groups lobby for regulations that could legally bind industries and consumers holding different priorities, how social media platforms sometimes enforce specific content rules upon billions of users globally, or how indigenous communities try to compel external developers to negotiate within the framework of traditional land tenure systems. These examples highlight how groups legitimately assert their norms in ways that directly impact outsiders.

What holds this complex patchwork together, then, isn't convergence on shared foundational values---which often remain deeply contested. Instead, social stability seems to rely on practical, workable mechanisms for managing disagreement non-violently: courts, legislatures, negotiation practices, and cultural habits of compromise \citep{mouffe1999deliberative, hampshire1999justice, ostrom2009understanding}. It's primarily when these conflict resolution processes falter that disruptive conflict threatens the quilt's integrity, not simply because disagreements exist. Our institutions and social norms, therefore, are best viewed not as reflections of timeless truths, but as pragmatic tools, developed through trial and error, to make coexistence possible amidst persistent diversity.

A key insight we can draw then is that what holds humans together despite profound disagreements is not value convergence, but practical mechanisms for coexistence---which we see as social technologies \citep{ostrom1990governing, hampshire1999justice, henrich2020weirdest}. These include the vast array of conventions, norms, and institutions that allow us to coordinate and cooperate, even when our underlying beliefs differ radically. Think of the intricate web of traffic laws enabling anonymous strangers to navigate cities safely despite differing destinations and driving skills; consider the diverse property rights regimes---from communal land tenure managed by village elders according to long-standing traditions, to complex intellectual property laws negotiated internationally---that regulate access to resources; or reflect on the formal procedures of legal systems, scientific peer review, and parliamentary debate, which provide structured ways to handle disputes and make collective decisions despite fundamental disagreements on substance. These social technologies are the stitches binding our patchwork quilt. Crucially, their effectiveness doesn't necessarily rely on universal buy-in or a majority's enthusiastic approval of their substance. Stability can emerge from perceived legitimacy (accepting a judge's unfavorable ruling because the process is deemed valid \citep{hadfield2014microfoundations}), coercion (obeying a disliked law backed by state power, like the federal enforcement of desegregation after Brown v. Board of Education) \citep{lessig1995regulation, sunstein1996social}, pragmatic self-interest (powerful stakeholders upholding a system they benefit from, even if others resent it \citep{marwell1993critical}), or simply the inertia of established practice \citep{gintis2010social}. A society can function even when many participants are merely behaving as-if they agree with its rules due to these pressures, while their internal disagreement persists \citep{bicchieri2005grammar}. The point is that these mechanisms facilitate coexistence; they don't demand uniformity of belief or value. They allow different patches of the quilt to maintain their distinct character while still being connected.

Of course not all disagreements are the same. Though we generally reject the fact/opinion distinction, we still distinguish between disagreements potentially resolvable through better information and deliberation (where the force of the better argument might help achieve convergence as in \cite{habermas1985theory}) and disagreements stemming from incompatible value systems or worldviews. For the latter, we have to shift from seeking agreement to managing conflict and enabling coexistence through shared practices and norms. This doesn't imply ``anything goes''. Often, the very frameworks enabling pluralistic coexistence rely on widely held---though still socially constructed and historically contingent---procedural norms or condemnations of certain extreme acts like war crimes \citep{sikkink2013justice}. These function not as timeless universal values, but as pragmatically defended principles within particular political systems (like liberal democracy) designed specifically to protect pluralism and prevent the breakdown of coexistence itself---even while their application remains contested and imperfect in practice \citep{finnemore1998international}.

This shift has profound implications for how we conceptualize AI safety and ethics. Instead of asking ``How do we align AI with human values?''---a question presupposing a single, coherent set of ``human values'' that can be discovered and encoded---we should ask the more fundamental question that humans have grappled with for millennia: \textbf{``How can we live together?''} This perspective embraces pluralism, contextualism, contingency, and the messy reality of human social life---the vibrant, diverse quilt we're all continually sewing together. Crucially, the two questions are not equivalent. ``How can we live together?'' acknowledges that a unified set of values cannot function as a practical foundation for society, and that the challenge is to build systems (both social and technical) that can navigate a landscape of deep and persistent disagreement.

\section{The Allure of the Clearer Vision: Alignment as Mistake Theory}

The patchwork quilt metaphor articulated in the quote from Richard Rorty that we included at the start of this article contrasts the pragmatic task of ``sewing together'' diverse elements with the ambition of ``getting a clearer vision of something true and deep''. While we advocate for the former, it's crucial to understand the deep appeal and implicit logic of the latter, particularly with regard to how it functions in the AI alignment community and its roots in rationalist and utilitarian traditions. Why might seeking this ``clearer vision'' seem not just desirable, but necessary for AI safety?

The framing resembles what Scott Alexander termed ``Mistake Theory'' \citep{alexander2018theory}. It is a perspective in which societal disagreements---even profound ones about values---are interpreted fundamentally as errors. They arise from cognitive biases, insufficient information, flawed reasoning, or incomplete deliberation. If these impediments could be overcome, perhaps through sufficiently advanced intelligence (human or artificial) applying universally valid principles of rationality and morality, convergence towards correct answers would be expected. There is, in this view, something ``true and deep''---a correct way to reason, a coherent underlying set of values (like a Coherent Extrapolated Volition \citep{yudkowsky2004coherent} or an optimal utility function), or a correct solution to coordination problems---that awaits discovery or clarification. Getting a ``clearer vision'' of this truth \emph{is} the path to progress and, crucially for AI safety, the path to ensuring reliably beneficial outcomes. If AI can be made to grasp this ``clearer vision'', its potentially vast intelligence could be anchored to something objectively correct, transcending the messy, error-prone disagreements of current human societies. The goal of alignment, then, becomes synonymous with enabling AI (and perhaps eventually humanity) to overcome its mistakes and perceive this underlying truth.

This perspective naturally informs many alignment approaches. Consider the pursuit of Coherent Extrapolated Volition \citep{yudkowsky2004coherent}. It posits that the outcome of an idealized process---imagining humans after extensive reflection, discussion, and resolving internal inconsistencies---defines the authoritative normative goal for AI. While sophisticated formulations acknowledge immense uncertainty about what this coherent volition would actually look like, and further put the uncertainty to work as a force that encourages caution and prevents the AI from overconfidently pursuing specific aims, the underlying philosophical move remains characteristic of Mistake Theory. It assumes that such an idealized, hypothetical state represents a truer or more correct reflection of human values than our current, messy, expressed preferences. The inconsistencies and disagreements we observe empirically are framed as mistakes or noise to be filtered out through the process of idealization. As Yudkowsky put it: ``The coherent extrapolated volition is our wish if we knew more, thought faster, were more the people we wished we were, had grown up farther together; [etc...]''~\citep{yudkowsky2004coherent}. Regardless of whether the coherent extrapolated volition is conceived as a concrete target or an unreachable theoretical limit, it is granted normative precedence. It is exactly this process of attempting to successively approximate the idealized but in practice unknowable coherent extrapolated volition that we identify with Rorty's phrase ``getting a clearer vision of something true and deep''. Examples include methodologies drawing on ideal observer/advisor theories (e.g.~\cite{firth1952ethical, muehlhauser2013ideal}) or aiming for reflective equilibrium (e.g.~\cite{rawls1971theory}) which seek to refine or correct flawed human judgments to arrive at a more rational, and implicitly more ``correct'', set of values or principles. The strong emphasis within parts of the alignment community on formal logic, decision theory, and long chains of abstract reasoning often carries an implicit assumption that these tools offer privileged access to objective truths, capable of overriding the seemingly less reliable outputs of culturally situated intuition or tradition. Even concerns about ``instrumental convergence'' \citep{bostrom2014superintelligence, turner2021optimal}---the idea that agents with diverse goals might converge on similar harmful subgoals (like power seeking)---often motivate the search for a single, robustly ``good'' objective function, grounded in that ``clearer vision'' of true value, to serve as a universal countermeasure. From the archetypal Mistake Theorist's standpoint, grounding AI safety in anything less than this pursuit of objective correctness seems fragile, susceptible to the very errors and disagreements it hopes to manage.

Our patchwork framework, in contrast, aligns more closely with ``Conflict Theory'' \citep{alexander2018theory}---though importantly, not in the specific political or conspiratorial terms Scott Alexander uses to frame it, but in the broader philosophical sense. This view takes disagreement seriously as a fundamental and often irreducible feature of the social landscape. Differences arise not just from correctable errors, but from genuinely divergent interests, incompatible values rooted in distinct ways of life, different historical experiences, and occupancy of different positions within social structures \citep{rorty1978philosophy}. It therefore resembles the views of \cite{mouffe1999deliberative} and \cite{march2011logic} which emphasize society's stability in the face of irreducible internal disagreement. From this perspective, expecting convergence on a single ``clearer vision'' is not only empirically dubious---given the persistence of deep disagreements throughout history despite intellectual progress---but potentially misguided. Progress, in this view, looks less like homing in on a preexisting Truth and more like the ongoing, difficult, practical work of ``sewing the quilt'': inventing, negotiating, and maintaining workable social arrangements, institutions, and norms that allow groups with fundamentally different outlooks to coexist, manage their conflicts non-destructively, and cooperate on shared practical goals despite deeper divisions. Morality and social order are seen as contingent human constructions \citep{rorty1989contingency}---practical solutions to the problem of living together---not reflections of a deeper, independent reality.

Why, then, do we favor the ``sewing'' over the ``seeing''? Because we believe the pursuit of the ``clearer vision'', while alluring in its promise of objective certainty, is ultimately less helpful for the practical task of building safe and beneficial AI within actual human societies. The persistence of pluralism suggests the ``single true vision'' may be a philosophical artifact rather than an achievable target \citep{mackie1977ethics}. Furthermore, we think the very attempt to identify and implement such a singular vision risks becoming coercive by implicitly or explicitly delegitimizing perspectives that don't fit the chosen model of rationality or value. Moreover, it distracts from the crucial, pragmatic work required: developing AI systems capable of navigating the existing landscape of disagreement, understanding context, respecting diverse norms, and participating constructively in the messy but vital human processes of negotiation, compromise, and conflict resolution.

\section{Potential Blind Spots in the Pursuit of Universal Alignment}

Reasoning and justification are controlled by epistemic norms \citep{kuhn1962structure, rorty1978philosophy}. Like all norms, epistemic norms vary from society to society and from subgroup to subgroup within a single society. Do arguments become convincing by appeal to repeatable empirical testing, rigorous logical deduction from axioms, adherence to sacred texts passed down through generations, deference to the wisdom of elders, or the consensus reached in communal deliberation? What counts as compelling evidence, legitimate argument, and sufficient justification varies dramatically, shaping everything from scientific inquiry and legal proceedings to how families make decisions, and how communities interpret misfortune. These observations, while not terribly controversial in themselves, have large implications for AI alignment research which have not often been articulated.

In this spirit, we have identified some common tendencies within the alignment paradigm, which while generally well-motivated, may inadvertently create blind spots or limitations worth examining. This isn't to dismiss the importance of rigor or the severity of potential risks, but to question whether approaches that implicitly downplay persistent disagreement and context-dependence are sufficiently robust.

\begin{enumerate}
    \item \textbf{The challenge of universalizing reasoning}: There's immense value in formal logic, probability theory, and cognitive bias reduction for improving clarity and truth-seeking. However, assuming there exists a single, optimal reasoning process applicable across all domains, which AI should ideally pursue, might be an oversimplification. Human justification and valid argumentation standards differ even between different parts of science \citep{kuhn1962structure, rorty1978philosophy}. While understanding the tendency to treat these differences as suboptimal deviations from an ideal Bayesian standard, we think it may be more productive to view them as locally effective epistemic \emph{norms} \citep{leibo2024theory}. They are locally appropriate standards for acceptance of justificatory argument. Ensuring powerful AI is robustly safe may require not just adherence to logical consistency, but also the capacity to navigate or translate between diverse, legitimate reasoning frameworks.

    \item \textbf{The fragility of long abstract arguments}: Exploring the logical consequences of assumptions through long chains of reasoning is a powerful tool, particularly for anticipating future possibilities or understanding complex systems like potential AGIs. However, relying heavily on such chains for safety assurances carries risks. Complex, abstract arguments can be highly sensitive to initial premises and definitions, where small divergences can lead to vastly different conclusions. Their length and abstraction may make them difficult to thoroughly verify, and prone to hidden errors or motivated reasoning \citep{mercier2017enigma, taber2016illusion}. As arguments extend further from empirical grounding, their reliability can decrease, even if each deductive step appears locally sound. Moreover, the well-known tendency for motivated reasoning to creep in, suggests that outsiders to any community will prudently take positions that make them less likely to accept arguments supported by longer chains of reasoning, and more so the longer they grow \citep{leibo2024theory}.

    \item \textbf{The challenge of navigating heterogeneous preferences while pursuing shared goals (like X-risk mitigation)}: Compelling arguments highlight the potentially overriding importance of mitigating existential risks from advanced AI. We acknowledge the weight of these arguments. However, even granting the paramount importance of preventing human extinction, achieving this goal in practice seems inescapably entangled with the challenges of establishing and sustaining collective action in the patchwork society. This involves overcoming at least two fundamental subproblems of collective action that we think the alignment community substantially underinvested in relative to their likely importance \citep{marwell1993critical, heckathorn1996dynamics}: \begin{enumerate}
        \item \textit{The start-up problem} \citep{koster2020model}: It is all well and good to say the goal is to ``prevent human extinction'' as an abstract slogan (as many in the AI alignment community do), but once you define it sufficiently clearly to make it a concrete goal with real implications, then it immediately forces choices that reflect particular values. What constitutes acceptable near-term costs? Which aspects of human life are essential to preserve? Whose vision of a flourishing future is to serve as the model? These aren't trivial details. Even seemingly universal goods like ``survival'' are embedded in thick cultural contexts that shape their meaning and priority (in fact many cultures prioritize sacred values above survival~e.g.~ \cite{ginges2007sacred}). In general, mobilizing global action and resources towards any specific AI safety strategy will inevitably confront deep-seated disagreements rooted in different values, priorities, and worldviews regarding the nature of AI risks, the efficacy or fairness of proposed initial strategies, and the equitable distribution of upfront costs and responsibilities. Absent sufficient initial buy-in and agreement on a common path forward, or if key actors hesitate fearing others won't join, promising strategies may never get off the ground or if they do may still lack the foundational support needed for momentum \citep{marwell1993critical, koster2020model}.
        \item \textit{The free-rider problem} \citep{vinitsky2023learning}: Even if a collective AI safety initiative is successfully launched, its long-term viability and effectiveness are threatened by the incentive for individual actors to benefit from the resulting global public good (reduced X-risk) without contributing their fair share to its ongoing provision and maintenance. Costly safety measures, research contributions, or adherence to burdensome standards can be shirked by rational actors if they believe others will carry the load, potentially leading to an under-provision of global AI safety. Resolving the free-rider problem may require the design of sociotechnical systems for incentivizing provision and sanctioning non-contribution within a heterogeneous and decentralized global landscape \citep{heckathorn1996dynamics, ostrom2010polycentric}.
    \end{enumerate}
    \item \textbf{Mistake theory can be very solipsistic:} The assumption that others are merely mistaken probably fosters a psychologically healthier internal locus of control for an individual since it's individually empowering to believe that one's arguments, if only framed well enough, could lead to change \citep{alexander2018theory}. However, individual-level adaptiveness does not imply utility as a basis for understanding or intervening in complex social systems. Adopting the mistake theory perspective may therefore constitute a form of methodological solipsism, where a beneficial stance on the individual psychological level is unhelpfully generalized to the domain of group-level policy and socio-technical system design. The challenge of ensuring AI safety is about group-level coordination, governance, and the stable integration of AI into diverse societies---arenas where persistent disagreement and conflict dynamics are often central features, not mere mistakes. Strategies that assume, even implicitly, that rational convergence is the primary mechanism for social order appear to underestimate the importance of institutions and norms for managing conflict. Consequently, an over-reliance on Mistake Theory in the AI alignment community risks promoting groupthink toward solutions that are naive regarding the realities of power, pluralism, and the need for robust conflict management mechanisms, as folks working in this area may be prone to mistaking their own (healthy!) individual-centric epistemological stance for a sound foundation on which to make group-level policy recommendations.
\end{enumerate}

A successful approach to X-risk mitigation likely cannot afford to treat preference heterogeneity as mere noise to be filtered out. It may need, paradoxically, to incorporate robust mechanisms for navigating disagreement, negotiation, and context-dependent judgment within the basic framework itself. Otherwise, the solutions generated from it risk being socially unstable and unable to command the broad and enduring cooperation they require for their successful implementation.

Having acknowledged the appeal and the pitfalls of seeking the single ``clearer vision'', we now turn to elaborating the alternative: understanding the rich complexity of the human patchwork and the practical social technologies that hold it together.

\section{The Stitches That Bind: Conventions, Sanctions, and Norms}

Conflict Theory---the view that social disagreements stem from genuine conflicts of interest rather than mere misunderstandings---offers a powerful framework for understanding how diverse societies function \citep{alexander2018theory}. While Scott Alexander's formulation of Conflict Theory emphasizes power dynamics where dominant groups maintain position through coercion, our version focuses on how social technologies serve as ``stitches'' binding diverse elements into functional wholes despite fundamental disagreements. In our model, stable societies distribute power across multiple centers of authority rather than concentrating it in dominant factions \citep{ostrom2010polycentric}. This polycentric structure creates the conditions for stable cooperation through productive tension between competing interests, as interdependent groups find pragmatic incentives for compromise despite underlying value differences. Understanding these stitching mechanisms is crucial for designing AI systems that can navigate pluralistic environments without assuming or requiring value alignment.

We emphasize that the stability of human society need not be conceptualized as dependent on shared values or unified goals. Instead, in our view, societies function as elaborate patchworks of different communities, traditions, and belief systems---each with its own patterns and colors, stitched together into a functioning whole. What allows these diverse and often internally inconsistent groups to function, to cooperate, and even to flourish is not a utopian convergence of values, but rather a shared, often tacit, though sometimes explicit, understanding of appropriateness \citep{leibo2024theory}. This concept, encompassing conventions, norms, and institutions, serves as the key social technology for navigating inherent disagreement and fostering collective action. Think of a functioning legislature: representatives from opposing parties may hold fundamentally different views on most issues, yet they can still (ideally) engage in debate, compromise, and pass laws. This doesn't require them to align their underlying values; it just requires them to behave appropriately by following the laws, precedents, and norms that govern their behavior in the legislative context \citep{march2011logic}.

When considering instrumental convergence and potential risks from advanced AI systems, particularly future Artificial Superintelligences, we acknowledge the potential for dangerous power-seeking behaviors \citep{turner2021optimal}. However, from our Conflict Theory perspective, the emergence of power-seeking incentives in complex systems with diverse agents (human or AI) is expected. The fundamental risk lies not merely in the motivation to seek power, but in the potential for any single entity or tightly-coupled coalition to successfully concentrate overwhelming power, thereby destabilizing the entire system \citep{narayanan2025ai}. Thus, while the concern about powerful AI acting against human interests is valid, we believe the most productive path forward involves shifting focus from solely trying to eliminate power-seeking drives within individual AIs or humans (a potentially intractable problem) towards designing robust, adaptable, and resilient governance structures. This includes developing polycentric institutions and effective monitoring and sanctioning mechanisms that can manage power dynamics and maintain stability within the broader socio-technical system, even when populated by highly capable agents with diverse motivations \citep{ostrom1990governing, ostrom2010polycentric}. That is, we think the challenge is less about perfecting the internal alignment of any single agent, and more about strengthening the institutional `stitches' that hold the entire societal quilt together, ensuring no single patch can grow so dominant as to threaten stability of the whole quilt.

Understanding appropriateness requires distinguishing between what \cite{walzer1994thick} calls ``thick'' and ``thin'' moralities. ``Thick'' morality refers to the rich, culturally-embedded moral systems specific to particular communities---detailed norms, values, and practices that resist universal translation. These thick moral systems include complex conceptions of virtue, duty, honor, or piety that derive meaning from their cultural context. By contrast, ``thin'' morality attempts to distill universal principles that apply across contexts---e.g.~abstract notions like harm prevention or fairness that theoretically transcend cultural boundaries. Each culture's thick morality represents a distinct patch in our polychrome social quilt, with practices and norms that cannot be reduced to universal principles without significant loss of meaning \citep{williams1985ethics}. This distinction has profound implications for AI development: attempts to encode supposedly neutral or thin ethical principles inevitably import assumptions from particular thick moral traditions. Even systems designed to represent minimal ethical constraints reflect the thick moral assumptions of their designers about what constitutes appropriate AI behavior. Recognizing this cultural embeddedness doesn't render appropriateness purely subjective; rather, it acknowledges that meaningful ethical guidance for AI requires engaging with the rich complexity of situated human values rather than seeking a decontextualized universal framework in hopes that it could please everyone once we find it.

\section{Navigating Context and Building a Pluralistic AI Ecosystem}

The concept of appropriateness forms the bedrock of human social interaction, governing conduct, speech, and behavior through both explicit rules and implicit understanding. For AI systems to integrate effectively into our social environments, they must navigate this same landscape of contextually determined appropriateness. When AI systems fail to respect these context-dependent norms, the result is not best understood as ``misalignment'' with universal human values, but rather as context-inappropriate behavior---speaking or acting in ways that violate the tacit and explicit rules of particular social environments \citep{leibo2024theory}.

A crucial difference between humans and current AI actors is how they utilize context. Humans know the stakeholders and nuances; AI models often receive only the bare text prompt. This lack of context can lead to behavior that tears at the social fabric simply because the system is unaware of who it's interacting with or why. To make this concrete, think about two applications: `corporate tech support assistant' AI and a `comedy writing assistant' AI. Each operates under a different implicit contract: one demands precision and professional decorum, the other might invite creative risk. Appropriateness is dictated by this context---the application's role, user's purpose, audience---not the underlying model's capabilities \citep{leibo2024theory}. Application domains demanding divergent senses of the appropriate represent distinct patches on the quilt. Crucially, tolerance for specific failures varies wildly. For instance, a mildly offensive joke, perhaps acceptable within limits from the comedy bot depending on the user's request, would be completely unacceptable and inappropriate coming from the tech support helper.

Many current limitations arise because AI systems simply cannot access the same contextual information as human users. This lack of context can lead to behavior that seems misaligned---not because the system has an alien objective, but because it cannot sense who it is interacting with or why. It's a data problem just as much as a reasoning or intelligence problem. Without knowing whether they are conversing with an individual in a quiet library or a lively group in a bustling casino, AI model designers are forced to choose defaults that are appropriately bland for the least forgiving setting. Such caution is understandable for early applications, yet it restricts the richness of interaction necessary for many real-world use cases. Defaulting to blandness is a symptom of the monolithic mindset. We think that a more pluralistic and decentralized ecosystem would offer a path beyond blandness, towards richer, more context-aware, and more useful AI in every niche \citep{leibo2024theory}.

This context-starved default to bland, corporate-speak to avoid offense isn't just aesthetically displeasing. It is often functionally inadequate, representing a retreat from the specificity required for meaningful interaction. It fails to provide the empathy needed in sensitive conversations, the creative spark desired in brainstorming, or the tailored precision required for expert consultation. A truly useful AI ecosystem needs a collection of specialists, each capable of navigating a different human context. Crucially, this is not a point about AI capability generality. We are saying that, even when we assume a single AI system is capable on its own of handling every one of the cognitive demands arising in any context, the actual requirement of having a single system that operates in every context would entail bland LLM-speak, and thereby prevent it from actually working as well in any context as a more specialized system with the exact same cognitive capabilities \citep{leibo2024theory}.

These concerns lead us to suggest that a decentralized ecosystem of contextually specialized systems would be better than trying to build a singular, universally appropriate AI. The monolithic approach to AI appropriateness---attempting to encode a single set of behavioral standards applicable across all contexts---inevitably produces systems that are either overly restrictive in some domains or inappropriately permissive in others. Therefore we propose an approach with the following interconnected design elements:

\begin{enumerate}
    \item \textbf{Contextual grounding}: AI systems require access to rich contextual information about their operational environments. This includes not only conversation history but also broader data on the current situation such as geographical region, social roles, current events, cultural practices, and even environmental factors like time of day or local holidays---that humans automatically consider while planning their behavior.
    \item \textbf{Community customization}: Different communities should be able to shape the norms governing AI systems they use, reflecting their specific values and practices while respecting broader societal constraints.
    \item \textbf{Continual adaptation}: AI systems must continually update their understanding of appropriate behavior through ongoing feedback. This requires moving beyond static training toward continuous learning systems that can adapt to evolving social norms just as humans do, with sanctioning (both positive and negative feedback) serving as the universal interface through which this learning occurs.
    \item \textbf{Polycentric governance}: Decision-making about AI appropriateness should be distributed across multiple, overlapping centers of authority---from individual users to app developers to platforms to regulatory bodies---each with defined roles and responsibilities. This mirrors the structure of human social governance, where appropriateness is determined at multiple scales simultaneously.
\end{enumerate}

One may ask what happens when AI systems become powerful enough to shape, manipulate, or simply ignore the feedback mechanisms themselves? The appropriateness framework doesn't naively assume that feedback mechanisms are invulnerable. It recognizes that feedback mechanisms are themselves socially constructed and maintained. The solution isn't to abandon feedback in favor of finding the One True Objective Function---it's to design robust, decentralized feedback mechanisms that become stronger, not weaker, in the face of attempts to manipulate them.
 
This perspective suggests we should aim for a technical architecture that supports specialization, decentralization, and contextualization. Note that it is still possible to achieve this using a generic foundation model as the initial template from which many downstream context- and application-specific models could branch off. In that case, the base model provider would maintain fundamental safety guardrails, while downstream developers would adapt their model's sense of appropriateness for their specific contexts.

One of these proposals requires additional comment. The idea of improving AI systems' responsiveness to context by providing them with more data (contextual grounding) may be a controversial way to solve this problem. The necessary data sources produce personally-identifiable information, so the usual imperative is to keep such data private. In accord with the contextual integrity framework of \cite{nissenbaum2004privacy}, we view privacy as the property that labels appropriateness of information flow. Therefore, pursuing context-aware AI must go hand-in-hand with developing and deploying strong technical and governance solutions. This includes exploring privacy-preserving machine learning, on-device data processing, secure enclaves for sensitive computations, clear data governance rules, and other techniques designed to uphold user control and ensure information flows remain appropriate.

\section{Collective Flourishing: The Ever-Growing Quilt}

``How can we live together?'' goes beyond simply achieving a stable state of peace with minimal conflict. It's about creating the conditions for collective flourishing, where our patchwork society can grow, adapt, and innovate together.

When individuals consistently adhere to shared concepts of appropriateness, cooperative resolutions to challenges become more stable. This stability frees up valuable resources and energy for innovation and progress. Appropriateness, by discouraging actions that lead to conflict and helping to quickly resolve conflicts that do occur, enables societies to focus collectively on adding new patches to our ever-growing quilt.

Some might worry that our patchwork approach embraces a troubling relativism, but this misunderstands the quilt we're describing. Just as a quilt's structural integrity depends on solid stitching techniques regardless of pattern diversity, our appropriateness framework recognizes that while the specific content of norms varies across contexts, the social technologies that facilitate coexistence---the mechanisms of learning, adaptation, and conflict resolution---can be studied with rigor and implemented with care. At its core, our approach focuses on preventing the instability that arises when conflict spirals out of control and threatens to tear apart the social fabric that binds diverse communities together. Rather than pursuing the philosopher's stone of a universal objective morality---an endeavor that has repeatedly fractured along cultural and historical lines---we advocate for strengthening the practical social technologies that allow diverse patches to coexist without requiring them to adopt identical patterns. This emphasis on constructing robust mechanisms for managing disagreement, rather than eliminating it, enables us to focus on mitigating the very real risk of destabilizing conflict cascades. This is not relativism; it's pragmatism in the face of value pluralism.

\section{The Astronomer and the Tailor: Seeing Clearly vs. Sewing Well}

The difference between the alignment paradigm, as we've described its underlying assumptions, and the appropriateness framework mirrors the contrast in the Rorty quote at the start of this article.

The alignment approach casts the AI safety researcher implicitly as an Astronomer, peering through the telescope of reason and computation, seeking to ``get a clearer vision of something true and deep''. The goal is to map the objective structure of human values (perhaps the coherent extrapolated volition of \cite{yudkowsky2004coherent}) or the universal principle of rationality, believed to exist behind the distortions of our current biases, disagreements, and limited understanding. Success lies in accurately discovering this underlying reality and encoding it, thereby aligning powerful AI with objective correctness; failure means misinterpreting the cosmic map of value, leading AI astray towards potentially catastrophic ends. This quest embodies the core hope of Mistake Theory \citep{alexander2018theory}: that a correct, underlying order exists and can, with sufficient clarity and rationality, be discerned and used to guide us toward better outcomes, resolving disagreements by revealing the errors that generated them.

The appropriateness paradigm, conversely, casts the researcher as a Tailor, engaged in the pragmatic work \cite{rorty2021pragmatism} described as ``sewing together a very large, elaborate, polychrome quilt''. The tailor works not by discovering a preexisting pattern in the heavens, but by skillfully handling the diverse materials at hand---the existing patchwork of conflicting norms, values, traditions, and interests found in human societies. The goal is practical and context-dependent: to cut, fit, adjust, and stitch these disparate pieces together using social technologies (conventions, norms, institutions), creating social arrangements (and their integrated AI systems) that are functional and adaptable, allowing for coexistence despite the inherent variation and tension in the fabric. This approach accepts irreducible difference not as noise to be filtered out in pursuit of a singular truth, but as aspects of the primary material that require skillful management. It aligns with Conflict Theory's premise that managing disagreement to prevent instability is the fundamental task \citep{alexander2018theory}. Morality and social order, in this view, are contingent human inventions---practical solutions we continually make and remake together through ongoing negotiation and adaptation, not objective truths we find.

These contrasting metaphors highlight fundamentally different orientations towards the problem of AI safety. The Astronomer seeks certainty, universality, and the objectively correct solution, driven by the fear of misaligning AI with the True nature of value. Their focus is on perfecting the vision---defining the objective function, refining the ideal reasoning process, etc---as the necessary foundation for safety. The Tailor, embracing contingency and context, focuses instead on the process and the tools of construction: the feedback loops, the negotiation mechanisms, the adaptable governance structures, the ability for AI to learn and adjust its behavior based on the specific ``stitches'' required in different social settings. Their fear is not primarily of failing to grasp a universal truth, but of creating systems (social or technical) that are too rigid, too brittle, that tear the social fabric by ignoring context, imposing uniformity, or failing to manage inevitable disagreements constructively. Choosing the tailor's path means accepting that our task is not to align AI with a potentially illusory cosmic order of values revealed through pure reason, but to ground AI within the practical, ever-evolving, context-dependent solutions humans devise for living together. It means focusing on the intricate work of social integration and conflict management, not the lonely pursuit of objective certainty.

The universe does not owe us coherence. Human values do not promise convergence. This isn't pessimism---it's recognizing the actual pattern of human history, where we've demonstrably managed to live together \textit{despite} fundamental disagreements, \textit{not} by resolving them. In this vein, we suggest that it would be best for the AI alignment community to draw more inspiration than it has thus far from the ways that multicultural groups, containing multiple differing thick moralities, can live together despite their differences.

{\small
\bibliographystyle{abbrvnat}
\nobibliography*
\bibliography{main}

\begin{thebibliography}{63}
\providecommand{\natexlab}[1]{#1}
\providecommand{\url}[1]{\texttt{#1}}
\expandafter\ifx\csname urlstyle\endcsname\relax
  \providecommand{\doi}[1]{doi: #1}\else
  \providecommand{\doi}{doi: \begingroup \urlstyle{rm}\Url}\fi

\bibitem[Alexander(2018)]{alexander2018theory}
S.~Alexander.
\newblock Conflict vs. mistake.
\newblock \emph{Slate Star Codex}, 2018.

\bibitem[Amodei et~al.(2016)Amodei, Olah, Steinhardt, Christiano, Schulman, and
  Man{\'e}]{amodei2016concrete}
D.~Amodei, C.~Olah, J.~Steinhardt, P.~Christiano, J.~Schulman, and D.~Man{\'e}.
\newblock Concrete problems in {AI} safety.
\newblock \emph{arXiv preprint arXiv:1606.06565}, 2016.

\bibitem[Bicchieri(2005)]{bicchieri2005grammar}
C.~Bicchieri.
\newblock \emph{The grammar of society: The nature and dynamics of social
  norms}.
\newblock Cambridge University Press, 2005.

\bibitem[Bostrom(2014)]{bostrom2014superintelligence}
N.~Bostrom.
\newblock \emph{Superintelligence: Paths, dangers, strategies}.
\newblock Oxford University Press, Oxford, 2014.

\bibitem[Boyd et~al.(2011)Boyd, Richerson, and Henrich]{boyd2011cultural}
R.~Boyd, P.~J. Richerson, and J.~Henrich.
\newblock The cultural niche: Why social learning is essential for human
  adaptation.
\newblock \emph{Proceedings of the National Academy of Sciences}, 108\penalty0
  (supplement\_2):\penalty0 10918--10925, 2011.

\bibitem[Derex et~al.(2019)Derex, Bonnefon, Boyd, and Mesoudi]{derex2019causal}
M.~Derex, J.-F. Bonnefon, R.~Boyd, and A.~Mesoudi.
\newblock Causal understanding is not necessary for the improvement of
  culturally evolving technology.
\newblock \emph{Nature human behaviour}, 3\penalty0 (5):\penalty0 446--452,
  2019.

\bibitem[Derex et~al.(2025)Derex, Bonnefon, Boyd, McElreath, and
  Mesoudi]{derex2025social}
M.~Derex, J.-F. Bonnefon, R.~Boyd, R.~McElreath, and A.~Mesoudi.
\newblock Social learning preserves both useful and useless theories by
  canalizing learners’ exploration.
\newblock \emph{Proceedings B}, 292\penalty0 (2039):\penalty0 20242499, 2025.

\bibitem[Finnemore and Sikkink(1998)]{finnemore1998international}
M.~Finnemore and K.~Sikkink.
\newblock International norm dynamics and political change.
\newblock \emph{International organization}, 52\penalty0 (4):\penalty0
  887--917, 1998.

\bibitem[Firth(1952)]{firth1952ethical}
R.~Firth.
\newblock Ethical absolutism and the ideal observer.
\newblock \emph{Philosophy and Phenomenological Research}, 12\penalty0
  (3):\penalty0 317--345, 1952.

\bibitem[Fiske(1992)]{fiske1992four}
A.~P. Fiske.
\newblock The four elementary forms of sociality: framework for a unified
  theory of social relations.
\newblock \emph{Psychological review}, 99\penalty0 (4):\penalty0 689, 1992.

\bibitem[Ginges et~al.(2007)Ginges, Atran, Medin, and
  Shikaki]{ginges2007sacred}
J.~Ginges, S.~Atran, D.~Medin, and K.~Shikaki.
\newblock Sacred bounds on rational resolution of violent political conflict.
\newblock \emph{Proceedings of the National Academy of Sciences}, 104\penalty0
  (18):\penalty0 7357--7360, 2007.

\bibitem[Gintis(2010)]{gintis2010social}
H.~Gintis.
\newblock Social norms as choreography.
\newblock \emph{Politics, Philosophy \& Economics}, 9\penalty0 (3):\penalty0
  251--264, 2010.

\bibitem[Graham et~al.(2009)Graham, Haidt, and Nosek]{graham2009liberals}
J.~Graham, J.~Haidt, and B.~A. Nosek.
\newblock Liberals and conservatives rely on different sets of moral
  foundations.
\newblock \emph{Journal of personality and social psychology}, 96\penalty0
  (5):\penalty0 1029, 2009.

\bibitem[Habermas(1985)]{habermas1985theory}
J.~Habermas.
\newblock \emph{The theory of communicative action: Volume 1: Reason and the
  rationalization of society}, volume~1.
\newblock Beacon press, 1985.

\bibitem[Hadfield and Weingast(2012)]{hadfield2012law}
G.~K. Hadfield and B.~R. Weingast.
\newblock What is law? a coordination model of the characteristics of legal
  order.
\newblock \emph{Journal of Legal Analysis}, 4\penalty0 (2):\penalty0 471--514,
  2012.

\bibitem[Hadfield and Weingast(2014)]{hadfield2014microfoundations}
G.~K. Hadfield and B.~R. Weingast.
\newblock Microfoundations of the rule of law.
\newblock \emph{Annual Review of Political Science}, 17:\penalty0 21--42, 2014.

\bibitem[Hadfield-Menell and Hadfield(2019)]{hadfield2019incomplete}
D.~Hadfield-Menell and G.~K. Hadfield.
\newblock Incomplete contracting and ai alignment.
\newblock In \emph{Proceedings of the 2019 AAAI/ACM Conference on AI, Ethics,
  and Society}, pages 417--422, 2019.

\bibitem[Hadfield-Menell et~al.(2019)Hadfield-Menell, Andrus, and
  Hadfield]{hadfield2019legible}
D.~Hadfield-Menell, M.~Andrus, and G.~Hadfield.
\newblock Legible normativity for ai alignment: The value of silly rules.
\newblock In \emph{Proceedings of the 2019 AAAI/ACM Conference on AI, Ethics,
  and Society}, pages 115--121, 2019.

\bibitem[Hampshire(1999)]{hampshire1999justice}
S.~Hampshire.
\newblock \emph{Justice Is Conflict}.
\newblock Princeton University Press, 1999.

\bibitem[Harris et~al.(2021)Harris, Boyd, and Wood]{harris2021role}
J.~A. Harris, R.~Boyd, and B.~M. Wood.
\newblock The role of causal knowledge in the evolution of traditional
  technology.
\newblock \emph{Current Biology}, 31\penalty0 (8):\penalty0 1798--1803, 2021.

\bibitem[Heckathorn(1996)]{heckathorn1996dynamics}
D.~D. Heckathorn.
\newblock The dynamics and dilemmas of collective action.
\newblock \emph{American sociological review}, pages 250--277, 1996.

\bibitem[Henrich(2020)]{henrich2020weirdest}
J.~Henrich.
\newblock \emph{The {WEIRDest} People in the World: How the West Became
  Psychologically Peculiar and Particularly Prosperous}.
\newblock Farrar, Straus and Giroux, 2020.

\bibitem[Henrich(2021)]{henrich2021cultural}
J.~Henrich.
\newblock Cultural evolution: Is causal inference the secret of our success?
\newblock \emph{Current Biology}, 31\penalty0 (8):\penalty0 R381--R383, 2021.

\bibitem[Henrich et~al.(2005)Henrich, Boyd, Bowles, Camerer, Fehr, Gintis,
  McElreath, Alvard, Barr, Ensminger, et~al.]{henrich2005economic}
J.~Henrich, R.~Boyd, S.~Bowles, C.~Camerer, E.~Fehr, H.~Gintis, R.~McElreath,
  M.~Alvard, A.~Barr, J.~Ensminger, et~al.
\newblock “economic man” in cross-cultural perspective: Behavioral
  experiments in 15 small-scale societies.
\newblock \emph{Behavioral and brain sciences}, 28\penalty0 (6):\penalty0
  795--815, 2005.

\bibitem[Heyes(2022)]{heyes2022rethinking}
C.~Heyes.
\newblock Rethinking norm psychology.
\newblock \emph{Perspectives on Psychological Science}, 2022.

\bibitem[Iyengar and Massey(2019)]{iyengar2019scientific}
S.~Iyengar and D.~S. Massey.
\newblock Scientific communication in a post-truth society.
\newblock \emph{Proceedings of the National Academy of Sciences}, 116\penalty0
  (16):\penalty0 7656--7661, 2019.

\bibitem[Jagiello et~al.(2022)Jagiello, Heyes, and
  Whitehouse]{jagiello2022tradition}
R.~Jagiello, C.~Heyes, and H.~Whitehouse.
\newblock Tradition and invention: The bifocal stance theory of cultural
  evolution.
\newblock \emph{Behavioral and Brain Sciences}, 45:\penalty0 e249, 2022.

\bibitem[K{\"o}ster et~al.(2020)K{\"o}ster, McKee, Everett, Weidinger, Isaac,
  Hughes, Du{\'e}{\~n}ez-Guzm{\'a}n, Graepel, Botvinick, and
  Leibo]{koster2020model}
R.~K{\"o}ster, K.~R. McKee, R.~Everett, L.~Weidinger, W.~S. Isaac, E.~Hughes,
  E.~A. Du{\'e}{\~n}ez-Guzm{\'a}n, T.~Graepel, M.~Botvinick, and J.~Z. Leibo.
\newblock Model-free conventions in multi-agent reinforcement learning with
  heterogeneous preferences.
\newblock \emph{arXiv preprint arXiv:2010.09054}, 2020.

\bibitem[K{\"o}ster et~al.(2022)K{\"o}ster, Hadfield-Menell, Everett,
  Weidinger, Hadfield, and Leibo]{koster2022spurious}
R.~K{\"o}ster, D.~Hadfield-Menell, R.~Everett, L.~Weidinger, G.~K. Hadfield,
  and J.~Z. Leibo.
\newblock Spurious normativity enhances learning of compliance and enforcement
  behavior in artificial agents.
\newblock \emph{Proceedings of the National Academy of Sciences}, 119\penalty0
  (3):\penalty0 e2106028118, 2022.

\bibitem[Kuhn(1962)]{kuhn1962structure}
T.~S. Kuhn.
\newblock \emph{The structure of scientific revolutions}.
\newblock University of Chicago press Chicago, 1962.

\bibitem[Leibo et~al.(2024)Leibo, Vezhnevets, Diaz, Agapiou, Cunningham,
  Sunehag, Haas, Koster, Du{\'e}{\~n}ez-Guzm{\'a}n, Isaac, Piliouras, Bileschi,
  Rahwan, and Osindero]{leibo2024theory}
J.~Z. Leibo, A.~S. Vezhnevets, M.~Diaz, J.~P. Agapiou, W.~A. Cunningham,
  P.~Sunehag, J.~Haas, R.~Koster, E.~A. Du{\'e}{\~n}ez-Guzm{\'a}n, W.~S. Isaac,
  G.~Piliouras, S.~M. Bileschi, I.~Rahwan, and S.~Osindero.
\newblock A theory of appropriateness with applications to generative
  artificial intelligence.
\newblock \emph{arXiv preprint arXiv:2412.19010}, 2024.

\bibitem[Lessig(1995)]{lessig1995regulation}
L.~Lessig.
\newblock The regulation of social meaning.
\newblock \emph{The University of Chicago Law Review}, 62\penalty0
  (3):\penalty0 943--1045, 1995.

\bibitem[Machery and Stich(2022)]{machery2022moral}
E.~Machery and S.~Stich.
\newblock {The Moral/Conventional Distinction}.
\newblock In E.~N. Zalta, editor, \emph{The {Stanford} Encyclopedia of
  Philosophy}. Metaphysics Research Lab, Stanford University, {S}ummer 2022
  edition, 2022.

\bibitem[Mackie(1977)]{mackie1977ethics}
J.~L. Mackie.
\newblock \emph{Ethics: Inventing Right and Wrong}.
\newblock Penguin, 1977.

\bibitem[March and Olsen(2011)]{march2011logic}
J.~G. March and J.~P. Olsen.
\newblock {The Logic of Appropriateness}.
\newblock In \emph{{The Oxford Handbook of Political Science}}. Oxford
  University Press, 2011.
\newblock \doi{10.1093/oxfordhb/9780199604456.013.0024}.

\bibitem[Marwell and Oliver(1993)]{marwell1993critical}
G.~Marwell and P.~Oliver.
\newblock \emph{The critical mass in collective action}.
\newblock Cambridge University Press, 1993.

\bibitem[Mercier and Sperber(2017)]{mercier2017enigma}
H.~Mercier and D.~Sperber.
\newblock \emph{The enigma of reason}.
\newblock Harvard University Press, 2017.

\bibitem[Merry(1988)]{merry1988legal}
S.~E. Merry.
\newblock Legal pluralism.
\newblock \emph{Law \& society review}, 22\penalty0 (5):\penalty0 869--896,
  1988.

\bibitem[Mouffe(1999)]{mouffe1999deliberative}
C.~Mouffe.
\newblock Deliberative democracy or agonistic pluralism?
\newblock \emph{Social research}, pages 745--758, 1999.

\bibitem[Muehlhauser and Williamson(2013)]{muehlhauser2013ideal}
L.~Muehlhauser and C.~Williamson.
\newblock Ideal advisor theories and personal cev.
\newblock \emph{Machine Intelligence Research Institute}, 2013.

\bibitem[Narayanan and Kapoor(2025)]{narayanan2025ai}
A.~Narayanan and S.~Kapoor.
\newblock {AI} as normal technology.
\newblock \emph{Knight First Amendment Institute. KnightColumbia.Org}, 2025.

\bibitem[Nisbett(2003)]{nisbett2003geography}
R.~Nisbett.
\newblock \emph{{The Geography of Thought: How Asians and Westerners Think
  Differently...and Why}}.
\newblock Free Press, 2003.

\bibitem[Nissenbaum(2004)]{nissenbaum2004privacy}
H.~Nissenbaum.
\newblock Privacy as contextual integrity.
\newblock \emph{Wash. L. Rev.}, 79:\penalty0 119, 2004.

\bibitem[North(1990)]{north1990institutions}
D.~C. North.
\newblock \emph{Institutions, institutional change and economic performance}.
\newblock Cambridge University Press, 1990.

\bibitem[Ostrom(1990)]{ostrom1990governing}
E.~Ostrom.
\newblock \emph{Governing the commons: The evolution of institutions for
  collective action}.
\newblock Cambridge university press, 1990.

\bibitem[Ostrom(2009)]{ostrom2009understanding}
E.~Ostrom.
\newblock \emph{Understanding institutional diversity}.
\newblock Princeton University Press, 2009.

\bibitem[Ostrom(2010)]{ostrom2010polycentric}
E.~Ostrom.
\newblock Polycentric systems for coping with collective action and global
  environmental change.
\newblock \emph{Global Environmental Change}, 20:\penalty0 550--557, 2010.

\bibitem[Rawls(1971)]{rawls1971theory}
J.~Rawls.
\newblock \emph{A Theory of Justice}.
\newblock Harvard University Press, 1971.

\bibitem[Rorty(1978/2009)]{rorty1978philosophy}
R.~Rorty.
\newblock \emph{Philosophy and the Mirror of Nature}.
\newblock Princeton university press, 1978/2009.

\bibitem[Rorty(1989)]{rorty1989contingency}
R.~Rorty.
\newblock \emph{Contingency, irony, and solidarity}.
\newblock Routledge, 1989.

\bibitem[Rorty(2021)]{rorty2021pragmatism}
R.~Rorty.
\newblock \emph{Pragmatism as Anti-authoritarianism}.
\newblock Harvard University Press, 2021.

\bibitem[Sikkink and Kim(2013)]{sikkink2013justice}
K.~Sikkink and H.~J. Kim.
\newblock The justice cascade: The origins and effectiveness of prosecutions of
  human rights violations.
\newblock \emph{Annual Review of Law and Social Science}, 9\penalty0
  (1):\penalty0 269--285, 2013.

\bibitem[Sunstein(1996)]{sunstein1996social}
C.~R. Sunstein.
\newblock Social norms and social roles.
\newblock \emph{Colum. L. Rev.}, 96:\penalty0 903, 1996.

\bibitem[Taber and Lodge(2016)]{taber2016illusion}
C.~S. Taber and M.~Lodge.
\newblock The illusion of choice in democratic politics: The unconscious impact
  of motivated political reasoning.
\newblock \emph{Political Psychology}, 37:\penalty0 61--85, 2016.

\bibitem[Turiel(1983)]{turiel1983development}
E.~Turiel.
\newblock \emph{The development of social knowledge: Morality and convention}.
\newblock Cambridge University Press, 1983.

\bibitem[Turner et~al.(2021)Turner, Smith, Shah, Critch, and
  Tadepalli]{turner2021optimal}
A.~M. Turner, L.~Smith, R.~Shah, A.~Critch, and P.~Tadepalli.
\newblock Optimal policies tend to seek power.
\newblock In \emph{Proceedings of the 35th International Conference on Neural
  Information Processing Systems}, pages 23063--23074, 2021.

\bibitem[Ullmann-Margalit(1977)]{ullmann1977emergence}
E.~Ullmann-Margalit.
\newblock \emph{The emergence of norms}.
\newblock OUP Oxford, 1977.

\bibitem[Vinitsky et~al.(2023)Vinitsky, K{\"o}ster, Agapiou,
  Du{\'e}{\~n}ez-Guzm{\'a}n, Vezhnevets, and Leibo]{vinitsky2023learning}
E.~Vinitsky, R.~K{\"o}ster, J.~P. Agapiou, E.~A. Du{\'e}{\~n}ez-Guzm{\'a}n,
  A.~S. Vezhnevets, and J.~Z. Leibo.
\newblock A learning agent that acquires social norms from public sanctions in
  decentralized multi-agent settings.
\newblock \emph{Collective Intelligence}, 2\penalty0 (2):\penalty0
  26339137231162025, 2023.

\bibitem[Walzer(1994)]{walzer1994thick}
M.~Walzer.
\newblock \emph{Thick and thin: Moral argument at home and abroad}.
\newblock University of Notre Dame Press, 1994.

\bibitem[Wardhaugh and Fuller(2021)]{wardhaugh2021introduction}
R.~Wardhaugh and J.~M. Fuller.
\newblock \emph{An introduction to sociolinguistics}.
\newblock John Wiley \& Sons, 2021.

\bibitem[Whitehouse(2021)]{whitehouse2021ritual}
H.~Whitehouse.
\newblock \emph{The ritual animal: Imitation and cohesion in the evolution of
  social complexity}.
\newblock Oxford University Press, 2021.

\bibitem[Williams(1985)]{williams1985ethics}
B.~Williams.
\newblock \emph{Ethics and the Limits of Philosophy}.
\newblock Routledge, 1985.

\bibitem[Yudkowsky(2004)]{yudkowsky2004coherent}
E.~Yudkowsky.
\newblock Coherent extrapolated volition.
\newblock \emph{Singularity Institute for Artificial Intelligence}, 2004.

\end{thebibliography}
}

\end{document}